# Hybrid Quantum-Classical Neural Network for Cloud-supported In-Vehicle Cyberattack Detection

Mhafuzul Islam[1], Mashrur Chowdhury[1]*, Zadid Khan[1], and Sakib Mahmud Khan[1]
[1] *Glenn Department of Civil Engineering, Clemson University, Clemson, South Carolina 29634*
* *Senior Member, IEEE*

**Abstract—** A classical computer works with ones and zeros, whereas a quantum computer uses ones, zeros, and superpositions of ones and zeros, which enables quantum computers to perform a vast number of calculations simultaneously compared to classical computers. In a cloud-supported cyber-physical system environment, running a machine learning application in quantum computers is often difficult, due to the existing limitations of the current quantum devices. However, with the combination of quantum-classical neural networks (NN), complex and high-dimensional features can be extracted by the classical NN to a reduced but more informative feature space to be processed by the existing quantum computers. In this study, we develop a hybrid quantum-classical NN to detect an amplitude shift cyber-attack on an in-vehicle control area network (CAN) dataset. We show that using the hybrid quantum classical NN, it is possible to achieve an attack detection accuracy of 94%, which is higher than a Long short-term memory (LSTM) NN (87%) or quantum NN alone (62%)

## I. INTRODUCTION

The decoherence and mechanical errors in quantum computers can make it harder for the existing quantum computers to learn the underlying data pattern, affecting the performance [1]. With the recent advancement of near-term quantum processors, it is possible to use a combination of classical and quantum computers to reduce errors. In a hybrid quantum-classical setup some computations are performed in quantum computers and some computations are performed in classical computers. Such a setup can be used in a cloud-based cyber-physical systems (CPS) environment, where a control area network (CAN) bus is connected to the cloud using a CAN logger attached to the OBD-II port of a vehicle. The CAN logger provides CAN bus data to the cloud to run multiple CPS applications in the cloud while meeting the delay requirements (e.g., data upload and download delay) of the vehicle's operation (Figure 1) [2][3]. In this letter, the hybrid quantum-classical cyberattack detection application will run in the cloud to detect a cyberattack on the in-vehicle CAN bus. We consider an amplitude shift cyberattack, where an attacker can compromise an electronic control unit (ECU) locally or remotely and can perform an amplitude shift attack on the in-vehicle CAN bus. As the amplitude shift attack changes the data field of a CAN frame randomly, the complex nature of the attack makes it difficult to detect this kind of attack. Studies showed that CAN bus used in existing vehicles do not have sufficient security features [4] [5], and the security can be

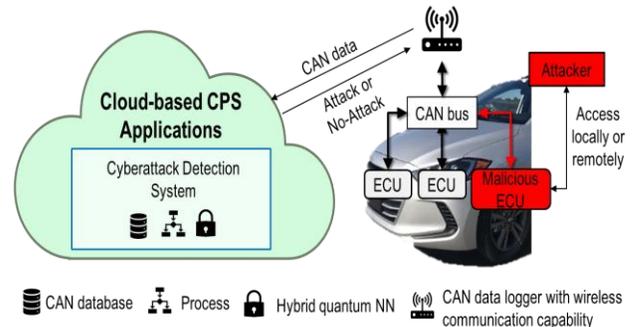

Fig. 1. Cloud-supported framework for in-vehicle cyberattack detection.

improved using machine learning techniques. The study by Song et al. shows an accuracy of 99% in detecting denial of service (DoS) attacks. However, their method will not work in the case of an amplitude shift attack, where the amplitude of a feature is shifted (up or down) randomly. The recent study by Khan et al. shows a detection accuracy of 87.8% on detecting amplitude shift attack using a deep neural network (DNN) [4]. To improve the attack detection accuracy, we combine a quantum machine learning method, more specifically a quantum NN, with a classical NN. By leveraging the advantages of the near-term quantum computers, the study by Farhi & Neven presented a general quantum NN architecture

---

Corresponding author: Mhafuzul Islam (mdmhafi@clemson.edu).

that was able to classify a handwritten digit dataset (i.e., MNIST) [6][7]. However, using such a quantum-only approach yields a lower classification accuracy. A more recent study shows the use of a hybrid quantum-classical NN approach can achieve a higher classification accuracy [8]. However, this approach has not been applied in a cloud-based in-vehicle cyberattack detection system. Using a cloud-based hybrid quantum-classical NN, we can overcome the existing limitations of quantum computers, and develop a quantum computing application for in-vehicle cyberattack detection. The objective of this study is to evaluate the performance of a cloud-supported hybrid quantum-classical neural network for amplitude shift attack compared to an existing LSTM NN or a quantum NN alone. The Section II and III discuss the amplitude shift attack model and the dataset used in this experiment. Different types of NN models for cyberattack detection are discussed in Section IV. Findings from our experiment are discussed in Section V. Finally, conclusions and future research directions are discussed in Section VI.

## II. ATTACK MODEL

We create the amplitude shift cyberattack following the study by Zadid et. al. [4]. In the amplitude shift attack, the amplitude of a feature of the in-vehicle network data is shifted (up or down) by a random constant value within a time interval. This random value can be both positive and negative, which is added to the original values of a feature. This simulates the scenario when an ECU is compromised by malware injection, which alters the course of ECU execution by adding or changing the amplitude of the output signal from an ECU. Although the amplitude of a feature changes in this attack, due to the addition of a constant value to the amplitude of a signal, the trend of variations over time remains unchanged. This complex cyberattack is more difficult to detect by traditional cyberattack detection system, as the trend remains same for the compromised feature.

## III. DATASET

To create an attack dataset, we first require an in-vehicle dataset which is attack-free. We have used one dataset created by the Hacking and Countermeasure Research Lab (HCRL) containing CAN data logged from a KIA soul vehicle [9]. The HCRL raw dataset contains different fields such as CAN ID, DATA and Timestamp. A generic DBC (Database CAN) file is used for decoding the raw CAN bus data. The DBC file is collected from the OpenDBC repository [10]. The decoding information in the DBC file is used to convert data bits into feature values. The features contain data from different in-vehicle sensors, as shown in Table I. The HCRL dataset contains 13 features with 95,200 timesteps (each timestep is 0.1s). These features correspond to different in-vehicle sensor values. We have used amplitude shift attack in this study. Using this attack, we create an attack dataset from the attack-free HCRL dataset. All features are compromised for one interval in the training dataset and another interval in the testing dataset. As a result, each feature is attacked once in training, and once in testing. The data from other features remain unchanged during attack on one feature. More details about the attack dataset can be found in Table II.

TABLE I
FEATURES IN HCRL DATASET

| Feature no. | Feature name | Value range |
|---|---|---|
| 1 | TQI_COR_STAT | 0.00-3.00 |
| 2 | TQI_ACOR | 0.00-99.61 |
| 3 | N | 0.00-16383.75 |
| 4 | TQI-EMS11 | 0.00-99.61 |
| 5 | TQFR | 0.00-99.61 |
| 6 | VS | 0.00-254.00 |
| 7 | BRAKE_ACT | 0.00-3.00 |
| 8 | TPS | 0.00-104.69 |
| 9 | PV_AV_CAN | 0.00-99.61 |
| 10 | TQI_MIN | 0.00-99.61 |
| 11 | TQI-EMS16 | 0.00-99.61 |
| 12 | TQI_TARGET | 0.00-99.61 |
| 13 | TQI_MAX | 0.00-99.61 |

## IV. CYBERATTACK DETECTION

TABLE II
DETAILS OF ATTACK DATASET

| Train / Test | Total no. of timesteps | Timesteps of attack on each feature | No. of total attack labels | No. of total non-attack labels |
|---|---|---|---|---|
| Train | 60000 | 2000 | 26000 | 34000 |
| Test | 35200 | 1000 | 13000 | 22200 |

In our hybrid quantum-classical NN (Figure 2), first, we preprocess the in-vehicle CAN bus dataset and construct a CAN image dataset [5]. Then we perform feature extraction using classical convolution neural network (CNN), convert the output from the classical CNN into quantum data, and use the quantum data into a quantum NN to detect an in-vehicle cyberattack. A measurement is performed from the quantum NN to obtain a classical output of attack or no attack.

### A. Data preprocessing

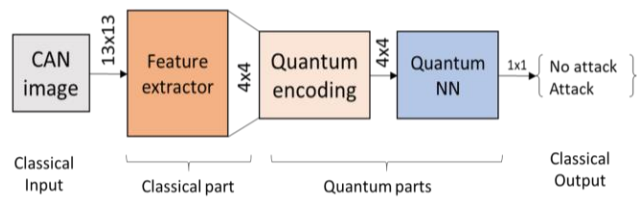

Fig. 2. Hybrid quantum-classical neural network.

We construct a $13 \times 13$ CAN image from 13 consecutive CAN frames, where each row represents a single CAN frame and each column represents a data feature. We consider a $13 \times 13$ CAN image as we have 13 data features in our dataset following the sillier methods presented in [4]. The study [5] used a CAN image size of 29x29 as the CAN id consist of 29 bits, whereas we only consider the 13 features of the CAN payload or message to construct a 13x13 image. The constructed CAN image dataset is represented by $D = \{(x_i, y_i)\}_{i=1}^{N}$, where $x_i$ is a $13 \times 13$ CAN image, with a label $y_i \in \{0,1\}$ representing no attack and attack image. $N$ is the number of total samples in $D$. We divide the total samples of, $N=7000$ into 80% training dataset (i.e., 5600 samples) and 20% testing dataset (i.e.,1400 samples). 21 The Table II shows the labeled CAN dataset where each CAN message represents one single data point. Here, as we take 13 features stacked up horizontally to create a 13x13 CAN image continuous CAN message stacked up vertically, and the total dataset size is reduced by a factor of 13 times.

*B. Feature extraction using classical neural network*

As presented in [5], we also use a CNN for extracting the features from a $13 \times 13$ CAN image and produce a $4 \times 4$ reduced image. The feature extraction from CNN can be represented as follows:

$$L_{4x4} = L_{n-1} \circ L_{n-2} \circ L_{n-3} \ldots L_{n1} \circ L_{n0} \qquad (2)$$

$$L_i: x_{i-1} \to x_i = \varphi(W_i x_{i-1} + b_i) \qquad (3)$$

Where, $L_{4x4}$ is the output of a CNN, $L_i$ is the $i^{th}$ layer of the CNN; $x_{i-1}$ and $x_i$ are the input and output vectors of $L_i$; $W_i$ is the weight, $b_i$ is a bias vector and $\varphi$ is a nonlinear function. Hyperparameters, such as number of layers $(n)$, $W_i$ and $b_i$ are optimized during the training phase on the training dataset for accurate classification.

*C. Quantum encoding*

To perform quantum operations (e.g., unitary operations, such as rotation, and phase flip of qubits), we need to convert the classical data into quantum data. We perform a quantum basis encoding to represent the classical data into quantum data [7]. In a quantum basis encoding, each encoded quantum state is the bit-wise translation of a classical binary data to the corresponding qubit of the quantum system. Here, each classical data is a N-bit binary string: $x^m = (b_1, \ldots b_N)$ with $b_i \in \{0,1\}$ for $i = 1, \ldots, N$. where, N features are represented with unit binary one bit, each input data $x^m$ can be directly mapped to the quantum state $|x^m\rangle$. Therefore, the entire dataset can be represented in superpositions of computational basis states as

$$|D\rangle = \frac{1}{\sqrt{M}} \sum_{m=1}^{M} |x^m\rangle \qquad (4)$$

Where, M is the total number of samples in a dataset, and each data sample can be represented as follows:

$$|x^m\rangle = \frac{1}{\sqrt{N}} \sum_{i=1}^{N} |b_i\rangle \qquad (5)$$

Where, N is the number of bits (i.e., 16 in our case), $b_i$ is a binary value represents if a feature is present or not, and $|b_i\rangle$ is the corresponding quantum qubit of the classical bit $b_i$. The binary image, $x^m$ is produced from $L_{4x4}$ using binary thresholding with a value of 0.5. Figure 3 shows the classical binary image before performing quantum encoding.

*D. Classification using quantum neural network*

With quantum encoded data, we train the parameterized quantum NN [6]. The parameterized quantum NN performs unitary operations, such as rotation, phase flip, on qubits and can be represented as follows:

$$Q = Q_{d-1} \circ Q_{d-2} \ldots Q_1 \circ Q_0 \qquad (5)$$

$$Q_m: |\varphi\rangle \to y = U(w)|x^m\rangle \qquad (6)$$

Where $Q$ is a binary output with $\{0,1\}$, where 0 and 1 represent no attack and attack detected, respectively; $Q$ has $d$ number of layers, $U(w)$ is a unitary operation on $|x^m\rangle$ with a weight $w$, and $y$ is the output after performing the unitary operation $U(w)$.

## V. EXPERIMENTAL SETUP

We compare the performance of the hybrid quantum classical NN with the quantum NN alone or LSTM based NNs. Each NNs gives an output to classify as attack (1) or no attack (0). For the quantum only NN, we have resized the 13x13 CAN image to 4x4 image, which is used as the input. The LSTM NN are developed following the study [4] having same model architecture. All NNs are developed using TensorFlow-quantum library available in google cloud platform [6]. We have executed our Hybrid quantum classical NNs and Quantum only NN models in the Cirq simulator provided by TensorFlow-quantum, and it represents an ideal noise-free quantum computer. We have performed the hyperparameter optimization such that we obtain the best cyberattack detection accuracy for each type of NNs. The optimized hyperparameters for each type of NNs are shown in the Table III.

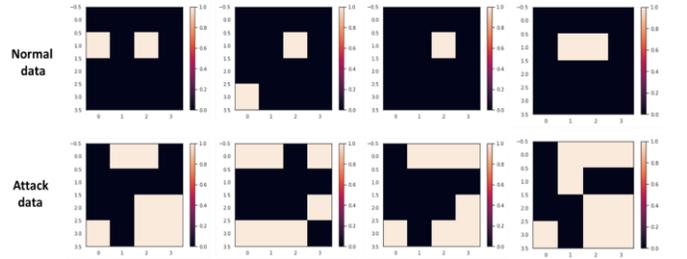

Fig. 3. Example of CAN feature image.

TABLE III
OPTIMIZED HYPERPARAMETERS OF DIFFERENT NEURAL NETWORKS (NNS)

| Hyperparameters | Hybrid Quantum-Classical NN | Quantum-only NN | LSTM NN |
|---|---|---|---|
| Number of qubits | 17 | 17 | N/A |
| Number of epochs | 20 | 50 | 100 |
| Number of layers | 6 | 7 | 3 |
| Batch size | 1 | 8 | 32 |
| Total trainable parameters | 96 | 128 | 1,090,049 |

## VI. EXPERIMENTAL RESULTS

As, the amplitude shift attack detection falls into a binary classification model (i.e., attack or no-attack), we used the classification accuracy as the performance metric with the following equation:

$$Accuracy = \frac{TP + FN}{TP + TN + FN + FP} \qquad (7)$$

Where TP = number of true positives, TN = number of true negatives, FP = number of false positives, and FN = number of false negatives. Figure 4 shows the attack detection accuracy on the training dataset and testing dataset. For both the training and testing dataset the hybrid quantum-classical NN shows 98.7% and 93.9% accuracy, respectively, whereas the quantum-only NN [7] shows 85.7%, and 62.0% accuracy on the training dataset and testing dataset, respectively. With the LSTM NN [4], the attack detection accuracy is 99.9% and 87.8% on the training and testing dataset, respectively. Here, the classical NN-based feature extractor was able to extract the features and the quantum neural network was able to perform more accurate attack detection. The feature map extracted from the classical NNs, CNN in this case, allowed the parameterized quantum neural network to explore the neighboring features in an exponentially large linear space, potentially allowing our hybrid-classical NN to capture the patterns in the dataset (i.e., statistical distributions) more efficiently than LSTM NN and quantum NN alone.

## VII. CONCLUSIONS

In a cloud-supported CPS environment, a hybrid-classical NN performs better in detecting an in-vehicle cyberattack compared to a quantum neural network, and a LSTM NN, as a hybrid-quantum NN, can capture the complex pattern of a cyberattack more efficiently. However, this study only demonstrates the use of quantum computers for amplitude shift cyberattack, which can be extended to detect other types of cyberattacks and computational performance will be conducted in future studies.

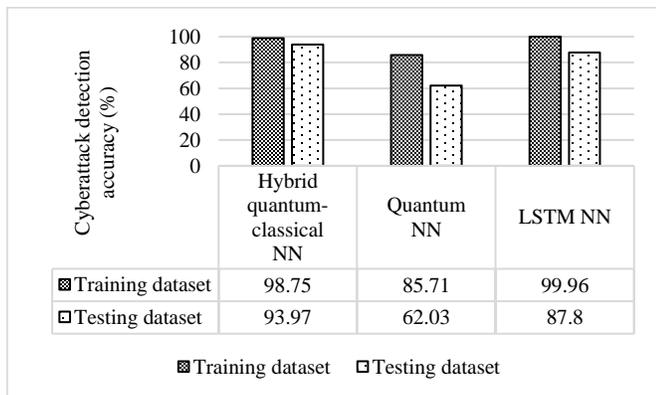

Fig. 4. Comparison of cyberattack detection accuracy for Hybrid quantum classical NN, Quantum NN and LSTM NN.

## ACKNOWLEDGMENT

This work is based upon the work supported by the Center for Connected Multimodal Mobility ($C^2M^2$) (a U.S. Department of Transportation Tier 1 University Transportation Center) headquartered at Clemson University, Clemson, South Carolina, USA. Any opinions, findings, conclusions, and recommendations expressed in this material are those of the author(s) and do not necessarily reflect the views of $C^2M^2$, and the U.S. Government assumes no liability for the contents or use thereof.